\documentclass[11pt]{article}

\usepackage[final]{acl}
\usepackage{times}
\usepackage{latexsym}
\usepackage[T1]{fontenc}
\usepackage[utf8]{inputenc}
\usepackage{microtype}
\usepackage{inconsolata}
\usepackage{graphicx}
\usepackage{amsmath}
\usepackage{amssymb} % \mathbb 포함
\usepackage{multirow}
\usepackage{algorithm}
\usepackage{algpseudocode}
\usepackage{placeins}
\usepackage{caption}
\algrenewcommand\algorithmicindent{1.2em}
\usepackage{array}  % p{..} + raggedright용
\usepackage{makecell}
\usepackage[table]{xcolor}
\usepackage{subcaption}
\usepackage{graphicx}

\usepackage{booktabs}
\usepackage{lipsum}

\title{Selective Test-Time Debiasing for CLIP via Reward Gating}

\author{
  Jaeho Han, Jisoo Yang, Hyeondong Woo, Mingyu Jeon, Sunjae Yoon, Junyeong Kim \\
  Department of Artificial Intelligence, Chung-Ang University \\
  \small\texttt{\{wogh50, yjs229, hyeondong, smart2557, sunjaeyoon, junyeongkim\}@cau.ac.kr}
}
\begin{document}
\maketitle

\begin{abstract}
Vision language models (VLMs) demonstrate strong zero-shot performance, but often perpetuate social stereotypes in person-centric queries, yielding skewed demographic distributions.
Current debiasing methods apply uniform bias corrections across all input queries regardless of their bias sensitivity, creating a fundamental fairness--utility trade-off.
Strong debiasing distorts semantically meaningful information in bias-insensitive queries, while weak debiasing fails to mitigate stereotypes in bias-sensitive ones.
This one-size-fits-all approach hampers simultaneously achieving high utility on bias-insensitive queries and fairness on bias-sensitive queries.
We introduce \textbf{Reward-Gated Test-Time Adaptation (RG-TTA)}, a reinforcement learning-based test-time adaptation framework that selectively applies debiasing based on input sensitivity.
RG-TTA adaptively triggers fairness regularization based on the bias sensitivity of each input during test-time policy adaptation, while focusing exclusively on optimizing cross-modal alignment for bias-insensitive inputs.
Experiments on fairness benchmarks (e.g., FairFace, UTKFace) demonstrate substantial bias reduction while simultaneously improving zero-shot utility, resolving the trade-off of uniform debiasing.
\end{abstract}

  % ===== Figure 1 =====

\begin{figure}[!t]
  \centering
  \includegraphics[width=\columnwidth]{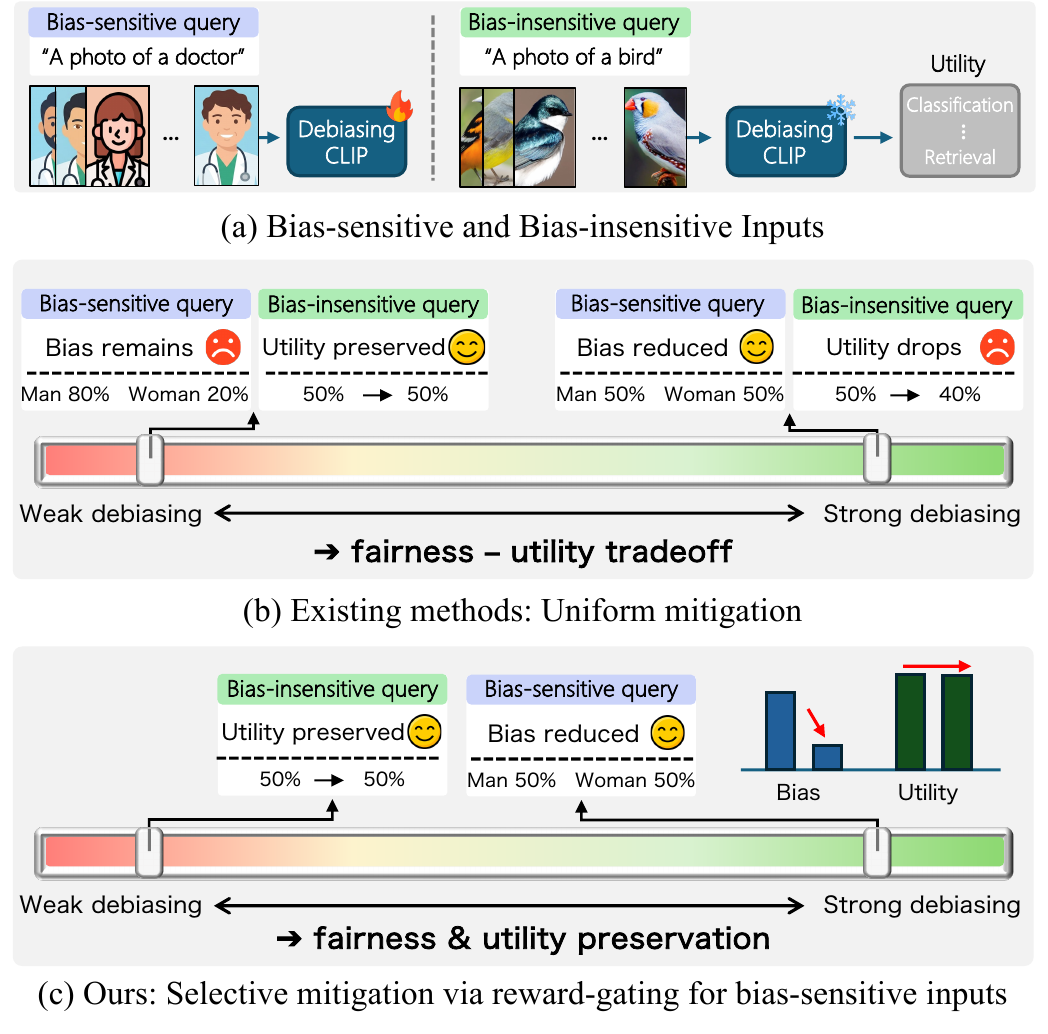}
    % \captionof{figure}{
    % \textbf{The fairness--utility trade-off in static debiasing.}
    % Existing methods apply a single global debiasing strength, resulting in
    % either insufficient bias mitigation or degraded utility.
    % }
    \captionof{figure}{
        (a) We categorize inputs into \textbf{bias-sensitive} and \textbf{bias-insensitive}, where only the former requires debiasing intervention.
        (b) Existing methods apply \textbf{uniform mitigation}, creating a structural trade-off: weak debiasing retains bias in sensitive queries (left), while strong debiasing distorts insensitive queries, degrading utility (right).
        (c) Our approach employs \textbf{selective mitigation via reward-gating}, which applies strong debiasing only to bias-sensitive inputs while preserving insensitive ones, ensuring both fairness and utility.
    }
\label{fig:teaser}
\end{figure}

\section{Introduction}

Vision Language Models (VLMs) have demonstrated exceptional zero-shot capabilities across a wide range of multimodal tasks~\citep{deng2009imagenet, plummer2015flickr30k}, reaching the stage of real-world applications. 
By learning joint representations from web-scale image-text pairs, these models achieve strong cross-modal alignment without task-specific fine-tuning.
However, this same training paradigm causes VLMs to internalize social stereotypes~\citep{birhane2021multimodal} present in their training data, leading to biased outputs that reflect and potentially amplify social prejudices~\citep{hall2023genderdisparities, hamidieh2024implicitbias, janghorbani2023multimodalbias, zhao2021racialbiascaptioning, wolfe2023contrastive, hausladen2025social}.
These biases manifest most critically in person-centric queries, where models produce skewed demographic distributions. 
For instance, querying ``a photo of a doctor'' yields disproportionately male images, or certain occupations become strongly associated with specific racial groups.
Such behavior poses serious risk of reinforcing discriminatory decision-making.

Existing debiasing approaches for VLMs~\citep{wang-etal-2021-gender, Chuang_2023_CVPR, Zhang_2025_CVPR} share common design philosophy: they apply fixed bias correction uniformly across all language queries, regardless of whether individual queries are sensitive to demographic biases.
Although conceptually simple, this uniform mitigation strategy causes a fundamental fairness--utility trade-off that we illustrated in Figure~\ref{fig:teaser}(b).
Here, we use utility to refer to general-purpose zero-shot performance on downstream tasks (e.g., image classification and cross-modal retrieval).

\begin{figure}[!t]    
    \centering
    \includegraphics[width=0.98\columnwidth]{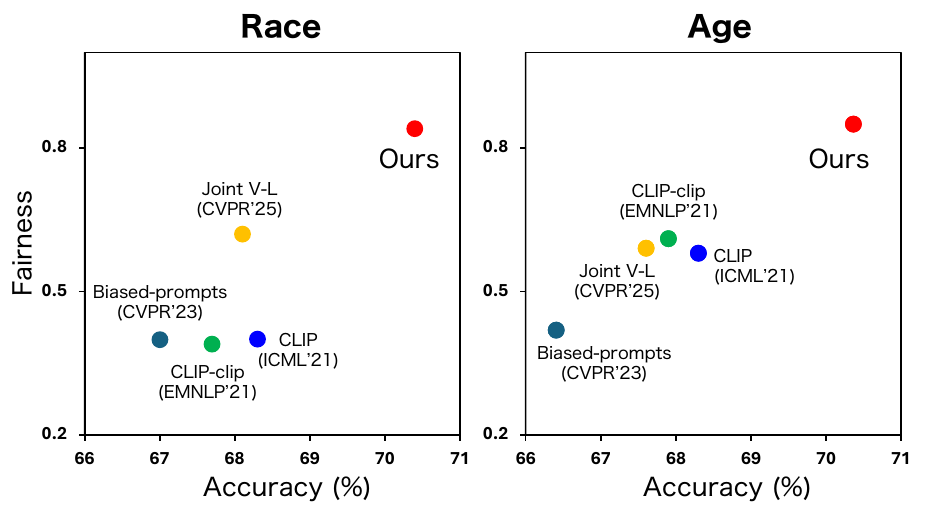}
    % \captionof{figure}{
    % Existing query-independent debiasing methods~\citep{Chuang_2023_CVPR, wang-etal-2021-gender, Zhang_2025_CVPR} exhibit a fairness to utility trade-off (i.e., Level of Fairness vs. ImageNet zero-shot Acc.), motivating our selective reward gating approach that achieves both lower bias and higher accuracy simultaneously.
    % }

\captionof{figure}{
Fairness versus utility for Race and Age.
\textbf{Accuracy} is measured as ImageNet zero-shot top-1 accuracy (\%),
and \textbf{Fairness} is measured as $1-\mathrm{MaxSkew@1000}$\protect\footnotemark\ (higher is better).
Existing query-independent debiasing baselines~\citep{Chuang_2023_CVPR, wang-etal-2021-gender, Zhang_2025_CVPR} exhibit a fairness--utility trade-off,
whereas our method improves fairness while achieving higher accuracy.
}
    \label{fig:visualization}
\end{figure}

For bias-sensitive queries, the model's predictions are often entangled with demographic attributes, and strong debiasing is necessary for fair outcomes.
In contrast for bias-insensitive queries, ground-truth semantics are largely orthogonal to demographic attributes, and the model's original predictions already reflect accurate cross-modal alignment.
When a uniform debiasing framework is applied to both types of queries, one of two failure modes inevitably occurs.
As we demonstrate in Figure~\ref{fig:visualization}, this inflexibility creates a trade-off where existing methods must compromise between fairness in sensitive queries and utility in general tasks, unable to excel at both simultaneously.
%

  % ===== Figure 2 =====
  % 글자크기 키우기
\protect\footnotetext{MaxSkew@1000 is computed from the protected-attribute distribution within the top-1000 retrieved samples for neutral queries; see Sec.~\ref{sec:method} for details.}

\begin{figure*}[t]
  \centering
  \includegraphics[width=\textwidth,trim=20 0 20
  .760,clip]{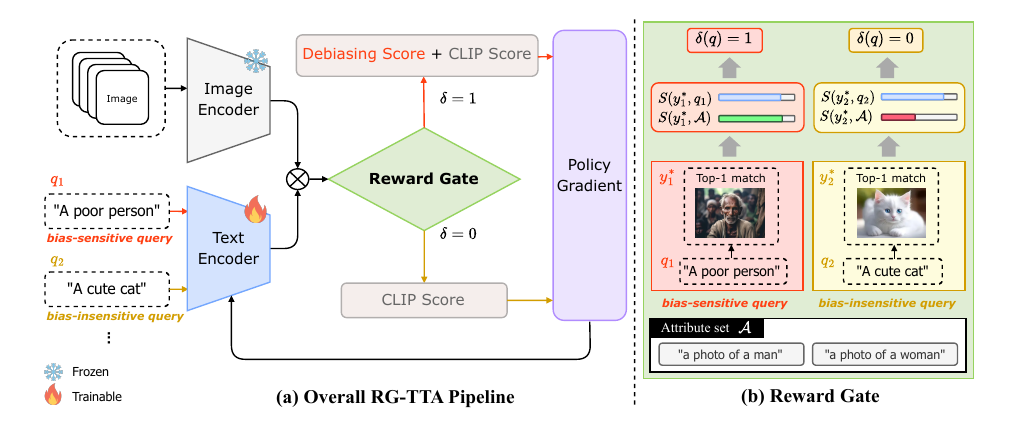}
\caption{Overview of RG-TTA: RL-based episodic test-time adaptation with \textbf{reward gating} via the indicator $\delta(q)$.
For each query, we update only the query-modality encoder (text encoder for text queries; image encoder for image queries) with a few policy-gradient steps on a truncated top-$K$ candidate set.
The indicator $\delta(q)$ controls the episode reward: when $\delta(q)=0$, we optimize an alignment-only reward; when $\delta(q)=1$, we add an attribute-balancing reward.
The gate activates when the top-1 match $y^*$ is sufficiently close to the attribute-alignment distribution, indicating elevated attribute entanglement.}

  \label{fig:figure3}
\end{figure*}

We believe that the aforementioned limitation is a consequence of the query-independent design paradigm.
Since uniform debiasing methods cannot distinguish between inputs that require mitigation and those that do not, they are constrained to operate in a compromise regime.
This observation motivates a paradigm shift toward adaptive debiasing that selectively activates mitigation based on the bias sensitivity of each input.
To this end, we propose \textbf{Reward-Gated Test-Time Adaptation for CLIP (RG-TTA)}, a reinforcement learning (RL)-based framework designed for selective debiasing. 
Our key insight is that debiasing should be treated as a per-query decision rather than global transformation, enabling the model to adapt its behavior dynamically based on input characteristics. 
As illustrated in Figure~\ref{fig:figure3}, RG-TTA operates through an episodic test-time adaptation protocol.
For each incoming query, we first assess its bias sensitivity by quantifying the alignment discrepancy between the query semantics and a set of demographic attributes.
Based on this assessment, an adaptive reward-gating strategy dynamically triggers the fairness-regularized objective only for bias-sensitive queries, ensuring the preservation of the original cross-modal alignment for neutral inputs.
Empirical evaluations on multiple fairness benchmarks--including FairFace~\citep{karkkainen2021fairface}, UTKFace~\citep{zhang2017utkface}, and the challenging FACET~\citep{gustafson2023facet} dataset--demonstrate that RG-TTA significantly reduces social bias across various demographic attributes.
Notably, our framework effectively resolves the fairness--utility trade-off by achieving substantial bias reduction alongside higher accuracy on tasks such as ImageNet-1K~\citep{deng2009imagenet} compared to existing query-independent baselines.
By performing episodic optimization with this gated reward, RG-TTA provides a practical design principle for mitigating bias without broadly disrupting the alignment of vision-language models.

%Instead of applying uniform mitigation, RG-TTA employs a selective reward-gating strategy that dynamically activates the fairness objective based on the estimated bias sensitivity of each input. As illustrated in Figure~\ref{fig:figure3}, the framework treats each test query as an independent episode: it selectively gates the fairness-regularized reward so that it is applied only to sensitive inputs, while leaving bias-insensitive queries unchanged to preserve their original cross-modal alignment. By optimizing this gated reward via a few REINFORCE-style policy-gradient steps, our method adapts the model at inference time and resets parameters immediately after each episode to prevent unnecessary drift. Empirically, RG-TTA significantly reduces social bias across diverse benchmarks such as FairFace and FACET while maintaining zero-shot utility on general tasks, effectively alleviating the structural fairness--utility trade-off.

% Our contributions are as follows:
% \begin{itemize}
%     \item \textbf{Adaptive Paradigm}: We propose \textit{selective debiasing} triggered by \textbf{bias-sensitive inputs}, addressing the structural limitations of uniform mitigation.
%     \item \textbf{RG-TTA Framework}: We introduce an RL-based framework using \textit{selective reward-gating} to dynamically balance fairness and utility.
%     \item \textbf{Empirical Validation}: We show across diverse benchmarks that RG-TTA alleviates the fairness--utility trade-off while \textbf{improving} zero-shot utility.
% \end{itemize}

%==========main figure=============

\section{Related Work}

\paragraph{Social debiasing in vision-language models.}
Social debiasing for CLIP-style VLMs is broadly categorized into (i) \emph{training-based} approaches~\citep{alabdulmohsin2024clipthebias, hirota2025biasgenderbenchmarks, Zhang_2025_CVPR}, which suppress sensitive-attribute signals by introducing additional objectives or modules during learning, and (ii) \emph{training-free} approaches~\citep{Chuang_2023_CVPR, gerych2024bendvlm}, which keep the foundation model fixed and apply post-hoc adjustments to embeddings or outputs. The former includes joint debiasing methods that align and remove biases across both modalities, while the latter estimates bias directions from attribute prompts and removes them via lightweight projections. However, most existing methods remain \emph{query-independent}, applying a single globally-defined transformation to all queries. A recent attempt toward query-adaptive debiasing is SANER~\citep{hirota2025saner}, which performs selective debiasing by restricting intervention to attribute-neutral text descriptions; yet its adjustment is confined to text features at training time, limiting adaptability to diverse input distributions encountered at deployment. Consequently, such fixed or partially selective rules still struggle to optimize for both objectives simultaneously, forcing a compromise between effective bias mitigation and the preservation of general model utility.

\paragraph{Test-Time Adaptation of CLIP via Reinforcement Learning.}
Test-time Adaptation (TTA) aims to improve model performance on unlabeled test inputs by performing parameter updates at inference time, enabling models to adapt to distribution shifts without retraining~\citep{sun2020ttt, wang2021tent}.
Early TTA methods relied on auxiliary self-supervised objectives such as entropy minimization~\citep{sun2020ttt, liu2021tttpp, wang2021tent}.
However, these approaches often suffer from instability from objective mismatch or prediction collapse~\citep{park2025retta}, motivating recent interest in utilizing VLM's internal alignment signal as direct feedback.
In particular, treating CLIP similarity as explicit reward and performing RL-based optimization~\citep{zancato2023retrievaltta} at inference time~\citep{zhao2024ttaclipreward} has been shown to reduce the collapse behavior observed in entropy-based updates and to improve zero-shot generalization.
%
% We build upon the feedback-driven TTA paradigm and extend it to address social bias in VLMs.
% %
% Specifically, we adapt CLIP via episodic RL updates with an input-dependent gating indicator that augments the alignment reward with a attribute-balancing reward only when needed, improving fairness without sacrificing general-purpose utility.
We build upon the feedback-driven TTA paradigm and extend it to address social bias in VLMs.
Our key contributions are (1) a bias-sensitivity gate that activates debiasing per input, and (2) a reward that combines cross-modal alignment with a bias-subspace debiasing signal.
With episodic test-time updates and selective reward gating, RG-TTA mitigates the fairness--utility trade-off, improving fairness and general-purpose utility.

\section{Method}
\label{sec:method}

We propose \textbf{Reward-Gated Test-Time Adaptation (RG-TTA)}, an RL-based framework for selective debiasing that adaptively updates model parameters during inference. RG-TTA is built on two key components. First, a \textbf{selective gating strategy} evaluates query sensitivity to trigger \textbf{fairness regularization} only when necessary. Second, an \textbf{adaptive reward function} balances CLIP alignment with an \textbf{attribute-balancing reward}, which encourages a uniform representation by favoring under-represented attributes. We update the model using a tractable approximation of the REINFORCE~\citep{williams1992simple} algorithm. By focusing updates on the most relevant candidates through top-$K$ truncation, this approach ensures that the optimization stays centered on query-specific semantics and limits drift via episodic resets.

\subsection{Preliminaries}
\label{sec:prelim}
\paragraph{CLIP.}
A pretrained vision-language model (VLM) consists of an image encoder $f(\cdot)$ and a text encoder $g(\cdot)$, which project both modalities into a shared embedding space~\citep{radford2021clip}.
Given a query $q$ and a candidate $y \in \mathcal{Y}$ selected from the opposite modality, we define an alignment score $S(q,y)$ as the cosine similarity between their embeddings.
Specifically, in the text-to-image setting we use $S(q,y)=\cos\!\big(g(q), f(y)\big)$, whereas in the image-to-text setting we use $S(q,y)=\cos\!\big(f(q), g(y)\big)$.
We use the policy score $S_\theta(q,y)$ for candidate ranking and parameterizing the policy. We use the reference score $S_{\mathrm{ref}}(q,y)$ only to compute the CLIP reward; $S_{\mathrm{ref}}$ is obtained from a fixed CLIP ViT-L/14 model throughout all experiments.

\paragraph{Test-time adaptation in vision--language tasks.}
Test-time adaptation (TTA~\citep{sun2020ttt, wang2021tent}) updates a trained model at inference time using a few unlabeled steps.
We follow an \emph{episodic} protocol: each query $q$ is adapted independently and the parameters are reset before the next query, which mitigates negative transfer.
For vision--language retrieval, where $q$ can be text or an image, we adapt only the query-modality encoder and keep the opposite-modality encoder fixed.

\subsection{Selective Gating Strategy}
\label{sec:gating}

% Applying debiasing uniformly to all queries is unnecessary and often induces undesirable parameter drift on bias-insensitive queries. To resolve the structural fairness--utility trade-off, we introduce a gating mechanism that evaluates the alignment discrepancy between query semantics and demographic attributes. This approach provides a signal to assess whether a query’s alignment is primarily supported by general semantics or shows elevated affinity to demographic attribute exemplars.

% For each query $q$, we first identify the top-1 candidate $y^*$ exhibiting the strongest CLIP similarity $S_\theta(q,y)$. As $y^*$ reflects the model's strongest match, we use it as an \textbf{anchor} to assess whether the alignment signal is influenced by demographic attribute associations. We then quantify the discrepancy by comparing $y^*$ against the mean attribute similarity computed over a predefined attribute set $\mathcal{A}$, which consists of representative exemplars for each protected group. Here, $\mathcal{A}$ is instantiated in the same modality as the query (e.g., attribute prompts for text queries and attribute exemplar images for image queries) so that $S_\theta(a, y^*)$ is computed using the same scoring function; concrete instantiations are provided in Sec.~\ref{sec:impl}.

Applying debiasing uniformly to all queries is unnecessary and can induce parameter drift on bias-insensitive queries. To mitigate the structural fairness--utility trade-off, we introduce a gating mechanism that measures the alignment discrepancy between query semantics and demographic attributes. For each query $q$, we select the top-1 candidate $y^*=\arg\max_y S_\theta(q,y)$ as an \textbf{anchor} and assess whether the query–candidate alignment is disproportionately driven by demographic associations by comparing $y^*$ to the mean attribute similarity over a predefined attribute set $\mathcal{A}$ of protected-group exemplars. We instantiate $\mathcal{A}$ in the same modality as the query (e.g., attribute prompts for text and exemplar images for vision), so that $S_\theta(a,y^*)$ uses the same scoring function; concrete instantiations are given in Sec.~\ref{sec:impl}.

The gating logic is based on the following intuition: a large discrepancy suggests that the match is driven by general, attribute-independent semantic alignment, in which case we deactivate fairness regularization ($\delta(q)=0$) and focus on refining standard cross-modal alignment to enhance utility. Conversely, if $y^*$ is close to the attribute alignment distribution (within a threshold $\epsilon$), the signal is likely entangled with a specific attribute category, triggering an attribute-balancing reward ($\delta(q)=1$). Accordingly, the gate is defined as:

\begin{equation}
\label{eq:saad}
\delta(q)=\mathbb{I}\Big[\, S(q,y^*)-\frac{1}{|A|}\sum_{a\in A} S(a,y^*) < \epsilon \Big].
\end{equation}

Here, $\mathbb{I}[\cdot]$ denotes the indicator function and $\epsilon$ is a fixed threshold ($0.02$) kept constant throughout the episode to prevent parameter drift.

\subsection{Adaptive Reward Function}
\label{sec:reward}

In this section, we define a candidate-wise reward $r(q,y_k)$ for each candidate $y_k \in \mathcal{Y}_K(q)$, where $\mathcal{Y}_K(q)$ denotes the candidate set used for adaptation. We describe the construction of $\mathcal{Y}_K(q)$ in Sec.~\ref{sec:opt}. Our reward always includes a CLIP-based alignment signal to preserve general cross-modal matching ability, and adds an attribute-balancing reward only when the gate is activated. This design maintains a balance between alignment and debiasing within each episode.

\paragraph{CLIP alignment reward.}
We follow prior work in defining a CLIP-based alignment reward~\citep{zhao2024ttaclipreward}. Concretely, for each candidate $y_k$ we compute a nonnegative alignment score as $s_k=\max\!\big(S_{\mathrm{ref}}(q,y_k),0\big)$, and use the episode mean $\bar{s}=\frac{1}{K}\sum_{j=1}^{K}s_j$ as a baseline. The resulting baseline-normalized alignment reward is

\begin{equation}
\label{eq:rclip}
r_{\text{clip}}(q,y_k)= s_k-\bar{s}.
\end{equation}

\paragraph{Bias subspace and attribute-balancing reward.}
When $\delta(q)=1$, we incorporate an attribute-balancing reward computed in a predefined \emph{bias subspace} into the reward. 
Let $\mu_c \in \mathbb{R}^{D}$ denote the class-mean embedding for class $c$ in the shared embedding space, precomputed from a labeled source dataset and kept fixed during test-time adaptation. 
We further define a reference vector $\bar{\mu}$, which can be instantiated as the average of $\{\mu_c\}_{c=1}^{C}$, and construct the set of difference vectors $\{\mu_c-\bar{\mu}\}_{c=1}^{C}$. 
We then perform PCA on these differences and extract all nonzero principal components, yielding $d$ components in total. 
The resulting orthonormal basis $U \in \mathbb{R}^{D \times d}$ defines the bias subspace\footnote{Projection-based techniques have also been used for group robustness on spurious correlations~\citep{zhu2025project}, in a different (training-time) setting and formulation.}

For each selected candidate $y_k$, let $z_k$ denote the embedding produced by the frozen encoder of the opposite modality. We then map both candidates and class prototypes into the bias subspace using the projection operator $P=UU^\top$, yielding $\tilde z_k = P(z_k-\bar{\mu})$ and $\tilde \mu_c = P(\mu_c-\bar{\mu})$. Next, we compute a soft assignment over classes by normalizing Gaussian-kernel similarities with a temperature parameter $\gamma$. Defining the squared distance between the projected candidate and class prototype as $d_{kc}=\|\tilde z_k-\tilde\mu_c\|_2^2$, the soft assignment is given by:

\begin{equation}
\alpha^{(k)}_c=
\frac{\exp(- d_{kc}/\gamma)}
{\sum_{c'=1}^{C}\exp(- d_{kc'}/\gamma)}.
\label{eq:soft-assign}
\end{equation}

To estimate the attribute distribution at the episode level, we compute the popularity for each class as $p_c=\frac{1}{K}\sum_{j=1}^{K}\alpha_c^{(j)}$. We then define the attribute-balancing reward for each candidate as
\begin{equation}
d(q,y_k)=\sum_{c=1}^{C}\alpha_c^{(k)}\!\left(p_c-\frac{1}{C}\right).
\end{equation}
% This score reflects how strongly candidate $y_k$ is associated with attribute classes that occur more or less often within the current episode, providing a query-conditioned signal that encourages a more balanced set of selected candidates.
This score reflects how strongly candidate $y_k$ is associated with attribute classes that are over- or under-represented in the current episode, encouraging a more balanced set of selected candidates.

\paragraph{Final combined reward.}
Finally, we define the overall reward for each selected candidate by combining the CLIP alignment reward with the query-conditioned attribute-balancing reward:
\begin{equation}
r(q,y_k)= r_{\text{clip}}(q,y_k) - \delta(q)\,\lambda\, d(q,y_k),
\label{eq:final_reward}
\end{equation}
where $\lambda$ is a hyperparameter that controls the strength of the attribute-balancing reward. When $\delta(q)=0$, the reward reduces to the CLIP alignment term alone. When $\delta(q)=1$, candidates associated with over-represented classes (i.e., $p_c > 1/C$) tend to have larger $d(q,y_k)$ and thus incur a larger balancing penalty, while candidates linked to under-represented classes (i.e., $p_c < 1/C$) tend to have smaller or negative $d(q,y_k)$ and are relatively favored. As a result, the episode-level attribute distribution is encouraged toward the uniform prior.

\subsection{Optimization and Episodic Update}
\label{sec:opt}

We optimize the proposed objective at test time via episodic policy-gradient updates~\citep{wang2021tent, shu2022testtime}, treating each input query $q$ as an episode.
We first evaluate the gating indicator $\delta(q)$ to determine whether to include the attribute-balancing term in the episode reward.
Conditioned on this decision, we choose a candidate budget $K$ (e.g., $K=10$ when $\delta(q)=0$ and $K=1024$ when $\delta(q)=1$), and construct a truncated candidate set $\mathcal{Y}_K(q)$ by selecting the top-$K$ candidates from the fixed pool $\mathcal{Y}$ according to the alignment score $S_\theta(q,y)$. We then compute the candidate-wise reward $r(q,y_k)$ for each $y_k \in \mathcal{Y}_K(q)$ and update the query-modality encoder parameters $\theta$ with a small number of gradient steps. After the episode ends, we reset the parameters to their initial state before processing the next query, mitigating negative transfer across queries.
We define the policy $\pi_\theta(y \mid q)$ over the full candidate pool $\mathcal{Y}$ using a softmax of the alignment score:

\begin{equation}
\pi_{\theta}(y \mid q)
=
\frac{\exp\!\big(S_{\theta}(q,y)\big)}
{\sum_{y' \in \mathcal{Y}} \exp\!\big(S_{\theta}(q,y')\big)} . 
\label{eq:policy}
\end{equation}
Although $\pi_\theta$ is normalized over the full candidate pool $\mathcal{Y}$ (i.e., the denominator is computed over $\mathcal{Y}$), we approximate the policy-gradient objective by summing only over the truncated set $\mathcal{Y}_K(q)$ for the current query. Concretely, for each episode we minimize the following REINFORCE-style objective:
\begin{equation}
\mathcal{L}(q)=-\frac{1}{K}\sum_{y_k\in\mathcal{Y}_K(q)} r(q,y_k)\,\log \pi_\theta(y_k \mid q).
\end{equation}
This top-$K$ truncation based on $S(q,y)$ yields a tractable approximation to the full policy-gradient objective while focusing updates on the most relevant candidates for the query.

\begin{table*}[t]
\tiny
\centering
\caption{Gender and age debiasing performance across different source datasets.
  We report results on in/out-of-domain fairness benchmarks and zero-shot utility tasks.
  ABLE is calculated based on in-domain metrics.
  Best results are shown in \textbf{bold}, and second-best results are \underline{underlined}.}
\label{tab:combined_results}

\setlength{\tabcolsep}{2.8pt}
\renewcommand{\arraystretch}{0.8}

\resizebox{\textwidth}{!}{
  \begin{tabular}{c|c|c|cc|cc|cc|cc|cc|c}
  \toprule
  \multirow{3}{*}{\centering Backbone} &
  \multirow{3}{*}{\centering Biases} &
  \multirow{3}{*}{\centering Methods} &
  \multicolumn{2}{c|}{In-Domain} &
  \multicolumn{4}{c|}{Out-of-Domain} &
  \multicolumn{2}{c|}{IN1K} &
  \multicolumn{2}{c|}{Flickr} &
  \multirow{3}{*}{\centering ABLE (\%)$\uparrow$} \\
  \cline{4-5} \cline{6-9} \cline{10-11} \cline{12-13}
  
  \rule{0pt}{2.0ex} & & &
  \multicolumn{2}{c|}{Source Dataset} &
  \multicolumn{2}{c|}{Cross Dataset} &
  \multicolumn{2}{c|}{FACET} &
  \multicolumn{2}{c|}{Acc. (\%)$\uparrow$} &
  \multicolumn{2}{c|}{R@5 (\%)$\uparrow$} &
  \\
  
  & & &
  {MS$\downarrow$} & {NDKL$\downarrow$} &
  {MS$\downarrow$} & {NDKL$\downarrow$} &
  {MS$\downarrow$} & {NDKL$\downarrow$} &
  {Top-1} & {Top-5} &
  {TR} & {IR} &
  \\
  \midrule

% ===================== UTKFace =====================
\rowcolor{gray!20}
\multicolumn{14}{c}{\textbf{Source: UTKFace | Cross: FairFace}} \\
\midrule

% ViT-B/16
\multirow{10}{*}{\centering ViT-B/16}
& \multirow{5}{*}{\centering Gender}
& Original CLIP   & 0.114 & 0.080 & 0.218 & 0.088 & 0.478 & 0.215 & \underline{68.31} & \underline{91.83} &  \underline{96.4} & 85.5 & 77.39 \\
& & CLIP-clip     & 0.070 & 0.055 & 0.133 & 0.038 & 0.459 & 0.190 & 67.81 & 91.42 &  95.4 & 83.0 & 78.52 \\
& & Biased-prompts & 0.179 & 0.062 & 0.161 & 0.048 & 0.460 & 0.215 & 65.07 & 89.38 &  94.3 & \underline{86.1} & 73.18 \\
& & Joint V-L     & \textbf{0.048} & \underline{0.043} & \underline{0.101} & \textbf{0.032} & \underline{0.456} & \underline{0.181} & 67.99 & 91.64 &  95.8 & 84.6 & \underline{79.36} \\
& & Ours          & \underline{0.051} & \textbf{0.029} & \textbf{0.080} & \underline{0.035} & \textbf{0.053} & \textbf{0.064} & \textbf{70.38} & \textbf{93.00} & \textbf{97.2} & \textbf{88.5} & \textbf{80.87} \\
\cmidrule(lr){2-14}
& \multirow{5}{*}{\centering Age}
& Original CLIP   & 0.421 & 0.229 & 0.657 & 0.433 & 0.744 & 0.367 & \underline{68.31} & \underline{91.83} & \underline{96.4} & \underline{85.5} &  66.96 \\
& & CLIP-clip     & \underline{0.393} & \textbf{0.215} & 0.643 & 0.430 & 0.745 & 0.364 & 67.93 & 91.58 & 96.1 & 84.5 &  \underline{67.70} \\
& & Biased-prompts & 0.578 & 0.451 & 0.777 & 0.550 & \textbf{0.635} & \underline{0.355} & 66.43 & 90.28 & 94.1 & 85.2 &  60.83 \\
& & Joint V-L     & 0.414 & 0.231 & \textbf{0.606} & \textbf{0.410} & 0.746 & 0.365 & 67.63 & 91.46 & 95.6 & 84.6 &  66.86 \\
& & Ours          & \textbf{0.151} & \underline{0.226} & \underline{0.641} & \underline{0.420} & \underline{0.742} & \textbf{0.330} & \textbf{70.36} & \textbf{92.99} & \textbf{97.2} & \textbf{88.3} & \textbf{77.39} \\
\midrule

% ViT-B/32
\multirow{10}{*}{\centering ViT-B/32}
& \multirow{5}{*}{\centering Gender}
& Original CLIP   & 0.066 & \textbf{0.032} & 0.138 & 0.054 & 0.485 & 0.225 & \underline{63.39} & \underline{88.83} & \underline{94.7} & \underline{83.5} & \underline{75.60} \\
& & CLIP-clip     & 0.098 & 0.045 & 0.253 & 0.105 & 0.500 & 0.240 & 62.21 & 88.23 & 93.0 & 81.0 & 73.79 \\
& & Biased-prompts & 0.089 & 0.036 & \underline{0.094} & \textbf{0.027} & \underline{0.417} & \underline{0.164} & 60.37 & 86.75 & 93.6 & 82.4 & 72.74 \\
& & Joint V-L     & \textbf{0.043} & \underline{0.033} & 0.108 & 0.039 & 0.469 & 0.212 & 62.46 & 88.23 & \underline{94.7} & 82.9 & \underline{75.60} \\
& & Ours          & \underline{0.054} & 0.038 & \textbf{0.088} & \underline{0.035} & \textbf{0.050} &  \textbf{0.021} & \textbf{69.76} & \textbf{92.31} & \textbf{97.0} & \textbf{86.6} & \textbf{79.60} \\
\cmidrule(lr){2-14}
& \multirow{5}{*}{\centering Age}
& Original CLIP   & 0.412 & \underline{0.253} & \underline{0.617} & \underline{0.416} & 0.752 & 0.388 & \underline{63.39} & \underline{88.83} & \underline{94.7} & \underline{83.5} & \underline{64.77} \\
& & CLIP-clip     & 0.415 & 0.264 & 0.659 & 0.435 & 0.754 & 0.397 & 62.70 & 88.31 & 94.2 & 83.2 & 64.34 \\
& & Biased-prompts & 0.522 & 0.409 & 0.701 & 0.497 & \textbf{0.663} & \textbf{0.366} & 61.07 & 86.92 & 92.0 & 82.2 & 60.19 \\
& & Joint V-L     & \underline{0.407} & \textbf{0.252} & 0.627 & \underline{0.416} & 0.751 & 0.370 & 62.93 & 88.66 & 94.1 & 82.5 & 64.69 \\
& & Ours          & \textbf{0.127} & 0.283 & \textbf{0.385} & \textbf{0.364} & \underline{0.741} & \underline{0.369} & \textbf{69.75} & \textbf{92.30} & \textbf{96.9} & \textbf{86.4} & \textbf{77.85} \\
\midrule

% ===================== FairFace =====================
\rowcolor{gray!20}
\multicolumn{14}{c}{\textbf{Source: FairFace | Cross: UTKFace}} \\
\midrule

% ViT-B/16
\multirow{10}{*}{\centering ViT-B/16}
& \multirow{5}{*}{\centering Gender}
& Original CLIP   & 0.218 & 0.088 & 0.114 & 0.080 & 0.478 & 0.215 & \underline{68.31} & \underline{91.83} &  96.4 & \underline{85.5} & 73.87 \\
& & CLIP-clip     & 0.103 & \underline{0.026} & 0.083 & 0.062 & 0.478 & 0.199 & 68.00 & 91.50 &  95.4 & 83.0 & 77.55 \\
& & Biased-prompts & 0.161 & 0.048 & 0.179 & 0.062 & 0.460 & 0.215 & 65.07 & 89.38 &  94.3 & 86.1 & 73.78 \\
& & Joint V-L     & \textbf{0.080} & \textbf{0.025} & \underline{0.040} & \underline{0.023} & \underline{0.446} & \underline{0.170} & 68.05 & 91.63 &  \underline{96.6} & 84.3 & \underline{78.35} \\
& & Ours          & \underline{0.082} & 0.031 & \textbf{0.030} & \textbf{0.022} & \textbf{0.114} & \textbf{0.040} & \textbf{70.32} & \textbf{92.98} & \textbf{97.2} & \textbf{88.5} & \textbf{79.75} \\
\cmidrule(lr){2-14}
& \multirow{5}{*}{\centering Age}
& Original CLIP   & 0.657 & 0.433 & 0.421 & 0.229 & 0.744 & 0.367 & 68.31 & \underline{91.83} & \underline{96.4} & \underline{85.5} &  58.94 \\
& & CLIP-clip     & 0.647 & 0.432 & 0.402 & \underline{0.215} & 0.742 & 0.373 & 67.97 & 91.61 & 96.3 & 84.4 &  59.16 \\
& & Biased-prompts & 0.777 & 0.550 & 0.578 & 0.451 & \textbf{0.635} & 0.355 & 66.43 & 90.28 & 94.1 & 85.2 &  54.33 \\
& & Joint V-L     & \underline{0.608} & \textbf{0.294} & \underline{0.377} & \textbf{0.115} & \underline{0.738} & \underline{0.341} & \underline{68.34} & 91.74 & 96.0 & 84.0 &  \underline{60.61} \\
& & Ours          & \textbf{0.526} & \underline{0.318} & \textbf{0.245} & 0.222 & 0.742 & \textbf{0.265} & \textbf{70.36} & \textbf{92.99} & \textbf{97.2} & \textbf{88.3} & \textbf{64.24} \\
\midrule

% ViT-B/32
\multirow{10}{*}{\centering ViT-B/32}
& \multirow{5}{*}{\centering Gender}
& Original CLIP   & 0.138 & 0.054 & 0.066 & 0.032 & 0.485 & 0.225 & \underline{63.39} & \underline{88.83} & 94.7 & \underline{83.5} & 73.37 \\
& & CLIP-clip     & 0.107 & 0.030 & \underline{0.061} & \underline{0.023} & 0.492 & 0.215 & 59.62 & 86.29 & 90.9 & 76.2 & 71.68 \\
& & Biased-prompts & 0.094 & \textbf{0.027} & 0.089 & 0.036 & \underline{0.417} & \underline{0.164} & 60.37 & 86.75 & 93.6 & 82.4 & 72.59 \\
& & Joint V-L     & \underline{0.090} & \underline{0.030} & \textbf{0.050} & \textbf{0.021} & 0.466 & 0.204 & 62.52 & 88.56 & \underline{94.9} & 82.9 & \underline{74.24} \\
& & Ours          & \textbf{0.078} & 0.032 & \textbf{0.050} & 0.029 & \textbf{0.137} &  \textbf{0.036} & \textbf{69.76} & \textbf{92.30} & \textbf{97.0} & \textbf{86.6} & \textbf{79.54} \\
\cmidrule(lr){2-14}
& \multirow{5}{*}{\centering Age}
& Original CLIP   & 0.617 & 0.416 & 0.412 & 0.253 & 0.752 & 0.388 & \underline{63.39} & \underline{88.83} & \underline{94.7} & \underline{83.5} & 58.29 \\
& & CLIP-clip     & 0.635 & 0.425 & 0.400 & 0.252 & 0.749 & 0.387 & 62.40 & 88.30 & 94.5 & 82.5 & 57.32 \\
& & Biased-prompts & 0.701 & 0.497 & 0.522 & 0.409 & \textbf{0.663} & \underline{0.366} & 61.07 & 86.92 & 92.0 & 82.2 & 54.76 \\
& & Joint V-L     & \underline{0.572} & \underline{0.364} & \underline{0.385} & \textbf{0.195} & 0.750 & 0.381 & 63.13 & 88.71 & 94.1 & 82.8 & \underline{59.60} \\
& & Ours          & \textbf{0.523} & \textbf{0.229} & \textbf{0.245} & \underline{0.222} & \underline{0.743} & \textbf{0.265} & \textbf{69.75} & \textbf{92.30} & \textbf{96.9} & \textbf{86.4} & \textbf{64.09} \\

\bottomrule
\end{tabular}
}
\end{table*}

\section{Experiments}
\label{sec:experiments}

\subsection{Datasets}
\label{sec:datasets}

To comprehensively evaluate both debiasing performance and generalization, we used a diverse set of benchmarks.
For fairness evaluation, we consider in-domain settings on FairFace (val)~\citep{karkkainen2021fairface} and UTKFace~\citep{zhang2017utkface}, and an out-of-domain setting on FACET~\citep{gustafson2023facet}.
To assess whether adaptation preserves general-purpose zero-shot utility, we additionally evaluate on ImageNet-1K~\citep{deng2009imagenet} for image classification and Flickr1k~\citep{plummer2015flickr30k} for retrieval.

\subsection{Metrics}
\label{sec:metrics}
\paragraph{Fairness metrics.}
Following prior work~\citep{berg-etal-2022-prompt, seth-etal-2023-dear}, we use retrieval-based metrics: MaxSkew@$k$ and NDKL@$k$. These metrics quantify the disparity in the distribution of protected attributes (e.g., gender, age, and race) within the top-k retrieved images for neutral queries. MaxSkew measures the maximum representation of a dominant group, while NDKL measures the divergence from uniform distribution. For both metrics, lower values indicate a fairer model.

\paragraph{Utility (V--L alignment) metrics.}
To ensure that our selective debiasing does not compromise the intrinsic V-L alignment of the pre-trained model, we evaluated zero-shot performance on standard benchmarks. We report Top-1 and Top-5 accuracy on ImageNet-1K\citep{deng2009imagenet} for classification, and Recall@5 for both Image-to-Text (TR) and Text-to-Image (IR) retrieval on Flickr1k\citep{plummer2015flickr30k}.

\paragraph{Alignment and Bias Level Evaluation (ABLE).}
Single metrics capture either fairness or utility, but not their trade-off. We use ABLE proposed by Zhang \textit{et al.}\citep{Zhang_2025_CVPR} for a holistic assessment. ABLE is defined as the harmonic mean of the zero-shot accuracy and the fairness score:
\begin{equation}
\mathrm{ABLE}
= \frac{2}{\frac{1}{acc} + \frac{1}{\exp(-\mathrm{MaxSkew}@k)}}
\end{equation}
where ${acc}$ denotes the ImageNet Top-1 accuracy. Higher ABLE indicates a better balance between mitigating social bias and retaining zero-shot accuracy; we report ABLE$\times$100 (\%) in tables.

%===============Table2 : UTKFace race성능==================
\begin{table}[t]
  \centering
  \caption{Race debiasing performance using UTKFace as the source.}
  \label{tab:src_utkface_race}

  \setlength{\tabcolsep}{3.5pt}
  \renewcommand{\arraystretch}{1.1}

  \resizebox{\columnwidth}{!}{
    % Backbone | Methods | UTKFace(2) | IN1K(2) | Flickr(2) | ABLE(1)
    \begin{tabular}{c|c|cc|cc|cc|c}
    \toprule
    \multirow{3}{*}{\centering Backbone} &
    \multirow{3}{*}{\centering Methods} &
    \multicolumn{2}{c|}{UTKFace} &
    \multicolumn{2}{c|}{IN1K} &
    \multicolumn{2}{c|}{Flickr} &
    \multirow{3}{*}{\centering ABLE (\%)$\uparrow$} \\
    
     \cline{3-4} \cline{5-6} \cline{7-8}
    
    & & & &
    \multicolumn{2}{c|}{\small Acc. (\%)$\uparrow$} &
    \multicolumn{2}{c|}{\small R@5 (\%)$\uparrow$} &
    \\
    
    & &
    {\footnotesize MS$\downarrow$} & {\footnotesize NDKL$\downarrow$} &
    {\footnotesize Top-1} & {\footnotesize Top-5} &
    {\footnotesize TR} & {\footnotesize IR} &
    \\
    \midrule

      % ===================== ViT-B/16 =====================
      \multirow{5}{*}{\centering ViT-B/16}
      & Original CLIP    & 0.575 & 0.137 & 68.31 & 91.83 & 96.4 & 85.5 & 63.31 \\
      & CLIP-clip        & 0.613 & 0.157 & 67.74 & 91.48 & 95.8 & 85.1 & 60.20 \\
      & Biased-prompts   & 0.604 & 0.208 & 67.00 & 90.72 & 94.1 & 85.8 & 60.21 \\
      & Joint V-L        & 0.378 & 0.069 & 68.07 & 91.64 & 96.5 & 83.8 & 68.30 \\
      & Ours             & \textbf{0.150} & \textbf{0.022} & \textbf{70.35} & \textbf{93.00}
                         & \textbf{97.2} & \textbf{88.3} & \textbf{77.42} \\
      \midrule

      % ===================== ViT-B/32 =====================
      \multirow{5}{*}{\centering ViT-B/32}
      & Original CLIP    & 0.698 & 0.213 & 63.39 & 88.83 & 94.7 & 83.5 & 55.75 \\
      & CLIP-clip        & 0.840 & 0.426 & 62.90 & 88.32 & 93.6 & 81.74 & 51.20 \\
      & Biased-prompts   & \textbf{0.317} & \textbf{0.140} & 61.80 & 87.46 & 91.9 & 83.34 & 60.21 \\
      & Joint V-L        & 0.638 & 0.230 & 63.01 & 88.56 & 94.6 & 82.4 & 57.48 \\
      & Ours             & 0.351 & 0.281 & \textbf{69.74} & \textbf{92.34}
                         & \textbf{97.0} & \textbf{86.5} & \textbf{70.07} \\
      \midrule
    \end{tabular}
  }
\end{table}
\subsection{Baselines}
\label{sec:baselines}

 % We compared our approach with the \textbf{Original CLIP} and three representative debiasing baselines across ViT-B/16 and ViT-B/32 backbones. \textbf{CLIP-clip}\citep{wang-etal-2021-gender} adopts a feature pruning strategy that utilizes the training dataset to estimate the mutual information between embedding dimensions and attribute labels, removing those with the highest bias correlation. \textbf{Biased-prompts}\citep{Chuang_2023_CVPR} is a training-free method that computes a projection matrix derived from biased text prompts to neutralize bias directions within the embedding space. Finally, \textbf{Joint V-L}\citep{Zhang_2025_CVPR} is a recent approach designed to mitigate the over-debiasing issue by jointly addressing biases in both image and text modalities. All methods were trained and evaluated on their ability to mitigate Gender and Age biases under consistent experimental settings.

We compare RG-TTA against the \textbf{Original CLIP} and three representative debiasing baselines using ViT-B/16 and ViT-B/32 backbones.
\textbf{CLIP-clip}~\citep{wang-etal-2021-gender} removes bias-correlated embedding dimensions identified via mutual information with attribute labels.
\textbf{Biased-prompts}~\citep{Chuang_2023_CVPR} neutralizes bias directions by projecting embeddings using prompt-derived attribute subspaces without retraining.
\textbf{Joint V-L}~\citep{Zhang_2025_CVPR} jointly debiases image and text representations to mitigate over-debiasing effects.
All methods are evaluated under identical settings for Gender, Age, and Race.
% \subsection{Implementation Details}
% \label{sec:impl}
% All experiments are conducted on an NVIDIA A6000 GPU with automatic mixed precision enabled.
% We use OpenAI-released pretrained CLIP checkpoints (ViT-B/16, ViT-B/32, and ViT-L/14), and compute the CLIP reward with a fixed CLIP ViT-L/14 model.
% During episodic TTA, we update only the encoder corresponding to the query modality: we adapt the text encoder for text-to-image retrieval, and adapt the image encoder for image-to-text retrieval and ImageNet classification; the opposite-modality encoder is kept frozen.
% We follow Sec.~\ref{sec:method} for the episodic TTA protocol and candidate construction.
% %
% We use AdamW~\citep{loshchilov2019decoupled} with learning rate $1\times10^{-4}$ and weight decay $0$.
% The reward gate uses a fixed threshold $\epsilon{=}0.02$. When $\delta(q){=}0$, we run $T{=}3$ update steps with $K{=}10$; when $\delta(q){=}1$, we run $T{=}10$ steps with $K{=}1024$, ranking candidates by CLIP alignment.
% We set $\lambda{=}1000$ and $\gamma{=}0.25$.
% For attribute-based gating, text prompts take the form “a photo of a \{attribute class\} person”, using one prompt per class. For image queries, we construct \(A\) by sampling \(M{=}5\) images per attribute class uniformly from UTKFace (fixed once and reused across all episodes), yielding \(|A| = C \times 5\).

\subsection{Implementation Details}
\label{sec:impl}
We use the official pretrained CLIP checkpoints~\citep{radford2021clip} with ViT-B/16, ViT-B/32, and ViT-L/14 backbones.
Consistent with the episodic TTA protocol, we update only the encoder corresponding to the input query modality (e.g., the text encoder for text-to-image retrieval) while keeping the target modality encoder frozen.
Optimization is performed using AdamW~\citep{loshchilov2019decoupled}.
Crucially, to balance efficiency and performance, we dynamically adjust the computational budget based on the gating decision $\delta(q)$: we perform a lightweight update with $T{=}3$ steps and $K{=}10$ candidates for bias-insensitive queries ($\delta(q){=}0$), while expanding to $T{=}10$ steps and $K{=}1024$ candidates for bias-sensitive ones ($\delta(q){=}1$).
Detailed hyperparameters, including learning rates, gating thresholds, and prompt templates, are provided in Appendix~\ref{sec:add_impl}.

% ============Figure 3: ablation study ============
\begin{figure}[t]
  \centering
  \includegraphics[width=\linewidth]{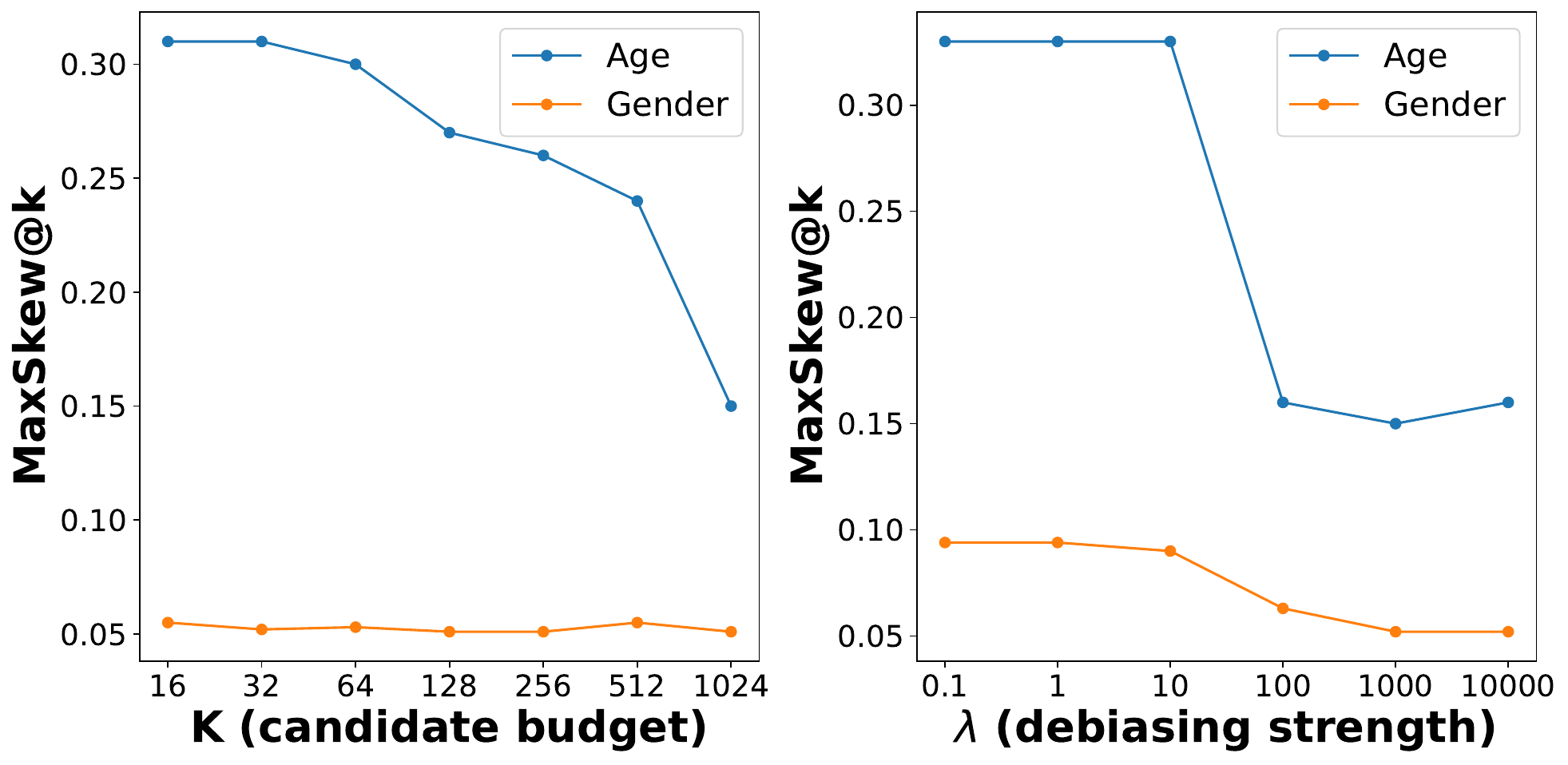}
\caption{Ablation of $K$ and $\lambda$ showing their effect on MaxSkew@$k$.}
  \label{fig:ablation_k_lambda}
\end{figure}

\subsection{Results}
Table~\ref{tab:combined_results} summarizes bias-mitigation results across different source datasets (UTKFace and FairFace) for Gender and Age. Overall, our method improves fairness in both in-domain and out-of-domain settings while enhancing ImageNet zero-shot performance. For Gender, we observe a clear in-domain improvement on UTKFace (MaxSkew@$k$: 0.114→0.051), with particularly large mitigation under distribution shift on FACET (MaxSkew@$k$: 0.478→0.053). Meanwhile, ImageNet Top-1 accuracy increases from 68.31 to 70.38, and ABLE also rises from 77.39 to 80.87, indicating that fairness gains do not come at the expense of utility (see Table~\ref{tab:combined_results} for full results).

This trend is consistent across attributes and source configurations: for Age and Race, we observe substantial in-domain fairness gains while maintaining (or slightly improving) zero-shot utility (Tables~\ref{tab:combined_results}, \ref{tab:src_utkface_race}). Switching the source dataset to FairFace shows the same behavior across in-domain and out-of-domain settings, aligning with our design—gating with episodic resets—that focuses updates on bias-sensitive queries while preventing drift on others.

% \FloatBarrier

%===============ablation study 2=====================
\begin{table}[t]
\centering
\small
\setlength{\tabcolsep}{5pt}
\renewcommand{\arraystretch}{1.12}
\caption{Ablation of reward variants for selective debiasing, evaluated on bias-sensitive (MS$\downarrow$) and bias-insensitive (IN1K Top-1$\uparrow$) benchmarks.}
\label{tab:ablation_qdg_core}
\begin{tabular}{lccc}
\toprule
Reward & MS$\downarrow$ & IN1K Top-1$\uparrow$ & ABLE (\%)$\uparrow$ \\
\midrule
CLIP          & 0.604 & 70.41 &  61.54\\
CLIP+Debias   & 0.149 & 53.98 &  58.23\\
RG-TTA(Ours)     & 0.151 & 70.36 & 77.39 \\
\bottomrule
\end{tabular}
\end{table}

\section{Discussion}
\subsection{Ablation Study}
We ablate key design choices to examine how selective reward-gating mitigates the fairness--utility trade-off.

\paragraph{Candidate budget and debiasing strength.}
Figure~\ref{fig:ablation_k_lambda} ablates the hyperparameters used when debiasing is active, varying the candidate budget $K$ and the debiasing strength $\lambda$ in the $\delta(q)=1$ regime.
We observe that increasing $K$ generally improves MaxSkew@1000 (notably for Age), suggesting that a larger truncated set yields a more stable estimate of the episode-level attribute distribution and thus a more reliable balancing signal.
Varying $\lambda$ shows a similar trend: weak values yield limited mitigation, while larger values deliver consistent gains with diminishing returns, suggesting performance is not brittle once $\lambda$ passes a minimal effective threshold.

%
% ============epsilon ablation===============
\begin{figure}[t]
  \centering
  \includegraphics[width=\linewidth]{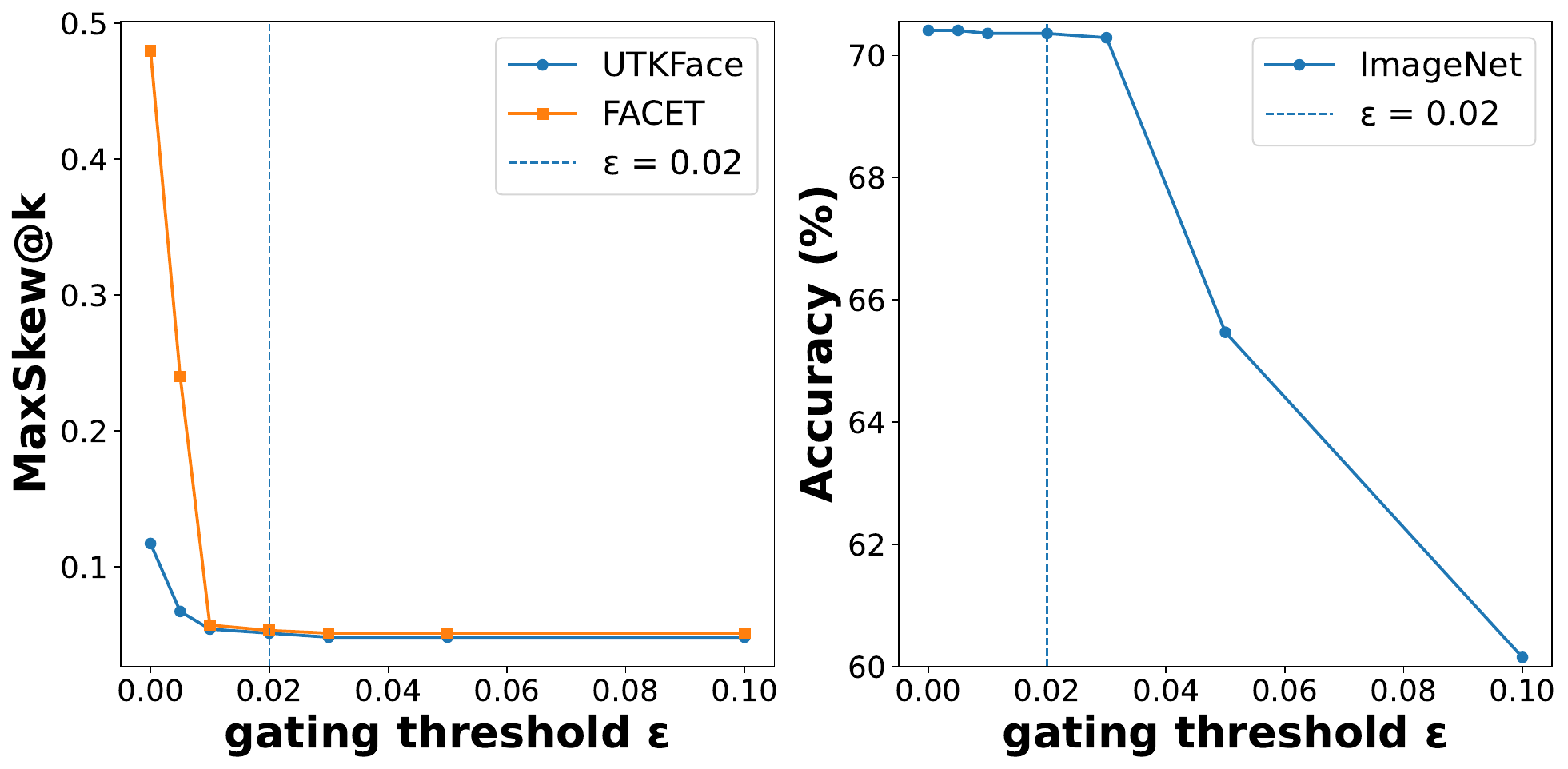}
\caption{Sensitivity to the gating threshold $\epsilon$. Left: MaxSkew@$k$ (gender) on UTKFace/FACET. Right: ImageNet-1K Top-1 accuracy.}
  \label{fig:ablation_epsilon}
\end{figure}

\paragraph{Selective vs.\ uniform debiasing.}
We next examine when to apply the debiasing signal versus how strongly to apply it.
Table~\ref{tab:ablation_qdg_core} compares (i) an alignment-only reward, (ii) adding the attribute-balancing term uniformly to all queries, and (iii) activating the balancing term only when the gating indicator $\delta(q)$ triggers.
While uniform debiasing can substantially reduce MaxSkew, it risks unnecessary adaptation and notably degrades zero-shot utility; in contrast, selective activation preserves utility while retaining the fairness gains, yielding a markedly better overall balance.

\paragraph{Gating threshold and activation behavior.}
We characterize the gate through threshold sensitivity and activation patterns under a dataset-level proxy setting.
Figure~\ref{fig:ablation_epsilon} shows that across a broad range of the threshold $\epsilon$, fairness improves rapidly and remains largely stable, and utility is stable around the default $\epsilon{=}0.02$, with noticeable degradation only when $\epsilon$ is overly large and activates debiasing too aggressively.
We also evaluate the gate's activation behavior (Table~\ref{tab:gate_fpfn}) under a proxy labeling scheme: treating UTKFace/FACET queries as debiasing-needed and ImageNet-1K queries as debiasing-not-needed, the gate activates for almost all UTKFace/FACET queries while remaining inactive for nearly all ImageNet-1K queries, yielding low false negative/positive rates in this proxy setting.

% Camera Ready : Reward model size start
\begin{table}[t]
\centering
\small
\caption{Same-size reward model ablation on UTKFace (Gender). ``Reward'' indicates the model used to compute the CLIP reward. Fairness improvements (MS) are preserved regardless of reward model size, while the larger ViT-L/14 reward primarily benefits utility (IN1K accuracy).}
\label{tab:same_size_reward}
\resizebox{\columnwidth}{!}{%
\begin{tabular}{llccc}
\toprule
Backbone & Reward & MS$\downarrow$ & IN1K$\uparrow$ & ABLE$\uparrow$ \\
\midrule
\multirow{3}{*}{ViT-B/16}
 & Original CLIP & 0.114 & 68.31 & 77.38 \\
 & RG-TTA (L/14) & 0.051 & 70.38 & 80.87 \\
 & RG-TTA (B/16) & 0.050 & 68.74 & 79.81 \\
\midrule
\multirow{3}{*}{ViT-B/32}
 & Original CLIP & 0.066 & 63.39 & 75.60 \\
 & RG-TTA (L/14) & 0.054 & 69.76 & 80.36 \\
 & RG-TTA (B/32) & 0.057 & 64.18 & 76.43 \\
\midrule
\multirow{3}{*}{ViT-L/14}
 & Original CLIP & 0.185 & 75.55 & 79.14 \\
 & RG-TTA (L/14) & 0.055 & 75.67 & 84.10 \\
\bottomrule
\end{tabular}%
}
\end{table}
\paragraph{Reward model scale.}
A natural concern is that RG-TTA relies on a larger reference model
(ViT-L/14) to compute the CLIP reward, which may be undesirable when
the backbone itself is smaller. We test whether fairness gains
depend on this scale mismatch by replacing the reward model with a
\emph{same-size} variant, matching the backbone
(Table~\ref{tab:same_size_reward}). The results reveal a clear pattern: \emph{fairness gains are
largely independent of reward model scale}. For ViT-B/16, the
same-size reward achieves MaxSkew of 0.050, essentially matching
the 0.051 obtained with the larger ViT-L/14 reward, and a similar
trend holds for ViT-B/32 (0.057 vs.\ 0.054). The benefit of the
larger reward model instead manifests in \emph{utility preservation}:
ImageNet accuracy improves more substantially when using ViT-L/14
as reward (e.g., B/16: 68.31$\rightarrow$70.38 vs.\ 68.74; B/32:
63.39$\rightarrow$69.76 vs.\ 64.18). Importantly, the same-size
configuration still improves ABLE over Original CLIP in all cases,
confirming that RG-TTA remains practically deployable when an
external larger reference model is unavailable, while the larger
reward model offers additional utility headroom when available.

% Camera Ready : Reward model size end

% Camera Ready : runtime analysis start

\subsection{Runtime Analysis}
\label{sec:runtime}

While RG-TTA introduces per-query test-time optimization,
understanding its computational cost is essential for assessing
practical deployment. We measure per-query wall-clock time on a
single NVIDIA A6000 GPU using ViT-B/16, averaged over 1000 queries.

\begin{table}[t]
\centering
\small
\caption{Per-query runtime on A6000 GPU with ViT-B/16.}
\label{tab:runtime}
\begin{tabular}{lc}
\toprule
Method & Inference/query \\
\midrule
Original CLIP & 4.49 ms \\
RG-TTA, $\delta(q){=}0$ & 110.96 ms \\
RG-TTA, $\delta(q){=}1$ & 467.41 ms \\
\bottomrule
\end{tabular}
\end{table}

\paragraph{Overhead analysis.}
As shown in Table~\ref{tab:runtime}, RG-TTA incurs additional
latency over Original CLIP due to alignment discrepancy computation
and policy gradient updates. The overhead ranges from
$\sim$25$\times$ for bias-insensitive queries ($\delta(q){=}0$) to
$\sim$104$\times$ for bias-sensitive queries ($\delta(q){=}1$),
reflecting the difference in candidate budget $K$ and update
steps $T$.

\paragraph{Practical implications.}
Despite this overhead, three factors support the practical
viability of RG-TTA. First, both conditions remain
\emph{sub-second}, which is acceptable for fairness-critical
batch-processing scenarios such as automated hiring systems or
medical image retrieval, where per-query latency of hundreds of
milliseconds is tolerable. Second, \emph{adaptive budgeting}
effectively reduces average overhead: as reported in
Table~\ref{tab:gate_fpfn}, the majority of general queries
are classified as bias-insensitive, so the average latency in
practice approaches $\sim$111\,ms rather than the worst-case
$\sim$467\,ms. Third, RG-TTA is entirely \emph{training-free};
unlike offline debiasing methods that require per-domain
fine-tuning, RG-TTA can be instantly applied to new
domains or attributes by replacing the bias subspace, offering
a favorable amortized cost in deployment scenarios with evolving
fairness requirements.

% Camera Ready : runtime analysis end

\subsection{Out-of-Domain Analysis on FACET}
On FACET (out-of-domain), our method transfers well for Gender, maintaining lower MaxSkew@$k$ than prior methods. We hypothesize that Gender is associated with multiple robust visual cues (e.g., face, hairstyle, clothing), which preserve group separation even after bias-subspace projection and lead to more stable debiasing signals. In contrast, Age relies on fine-grained facial cues that are sensitive to distance, resolution, and occlusion; in FACET's unconstrained images, these cues are weakened, reducing the reliability of the debiasing signal and ultimately limiting gains.

% ============ablation epsilon===============
\begin{table}[t]
  \centering
\caption{FP/FN are computed under a proxy labeling scheme where UTKFace and FACET are treated as positives (debiasing-needed) and ImageNet-1K as negatives (debiasing-not-needed).}
  \label{tab:gate_fpfn}
  \footnotesize
  \setlength{\tabcolsep}{4pt}
  \renewcommand{\arraystretch}{1.15}
  \begin{tabular}{lccc}
    \toprule
    Dataset & $\delta=1$ (\%) & $\delta=0$ (\%) & FP/FN \\
    \midrule
    UTKFace  & 99.9 & 0.1 & FN = 0.1 \\
    FACET    & 99.8 & 0.2 & FN = 0.2 \\
    ImageNet-1K      & 0.1 & 99.9 & FP = 0.1 \\
    \bottomrule
  \end{tabular}
\end{table}

\begin{figure}[t]
    \centering
    \setlength{\fboxsep}{0pt}
    \begin{subfigure}[t]{0.49\linewidth}
        \centering
        \fbox{\includegraphics[width=\linewidth]{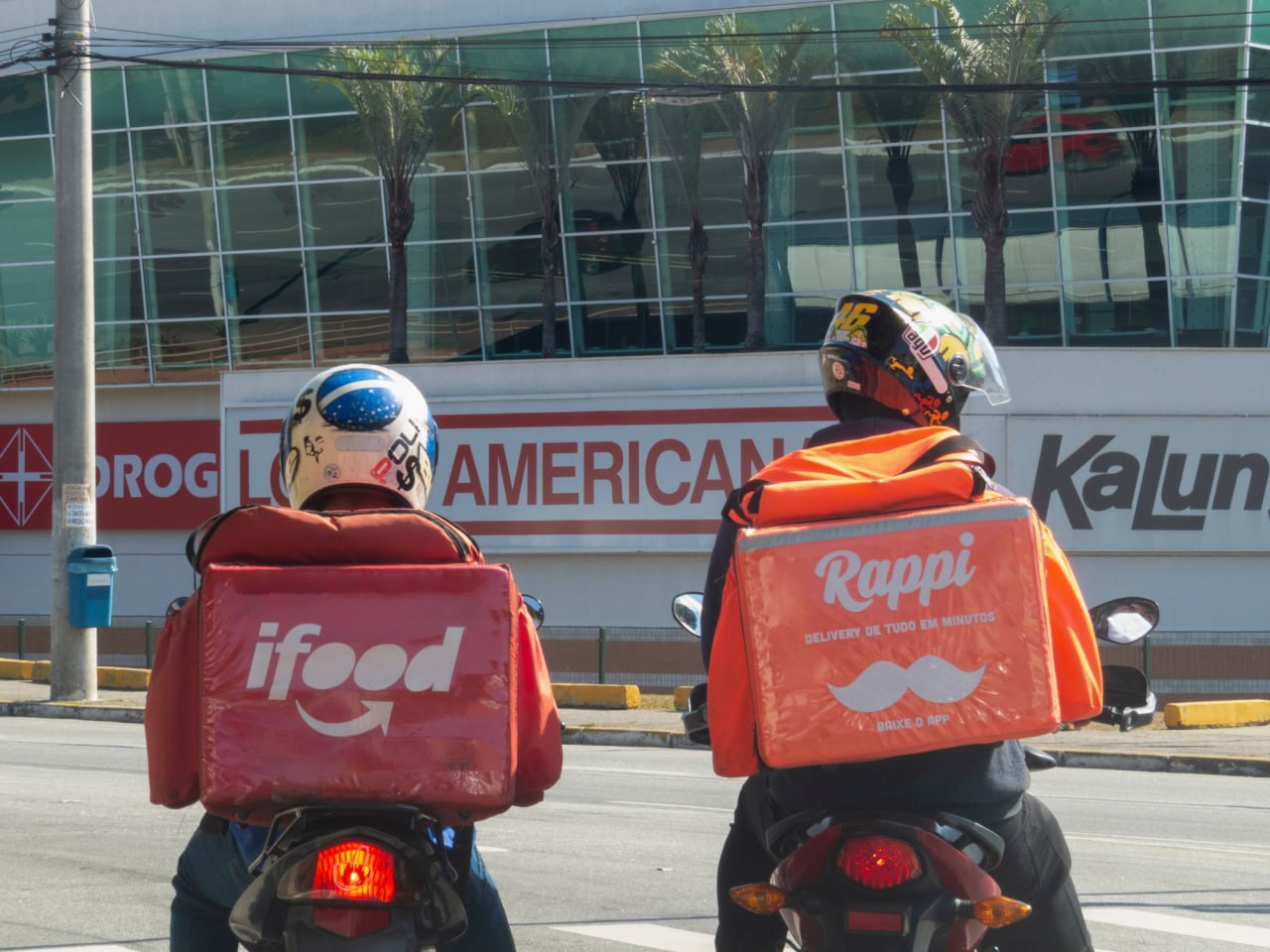}}
    \end{subfigure}\hfill
    \begin{subfigure}[t]{0.49\linewidth}
        \centering
        \fbox{\includegraphics[width=\linewidth]{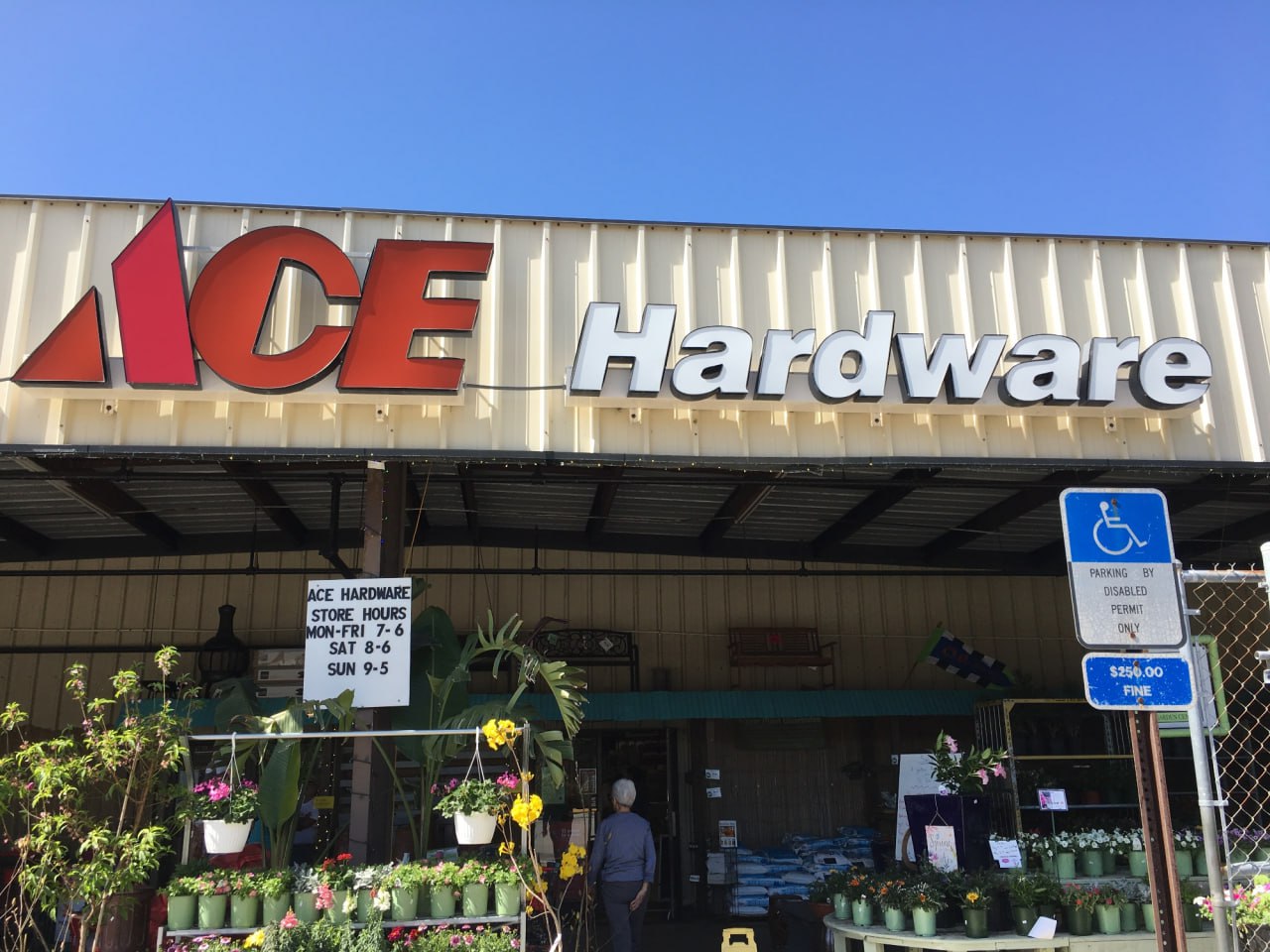}}
    \end{subfigure}
\caption{\textbf{Gating failure cases on FACET.} Bias-sensitive queries are incorrectly gated off ($\delta(q)=0$), preventing fairness regularization from being activated.}
    \label{fig:facet_gating_failures}
\end{figure}

\subsection{Gating Failure Case Analysis}
Figure~\ref{fig:facet_gating_failures} shows \emph{false-negative} cases on FACET, where bias-sensitive queries are gated off ($\delta(q)=0$) and thus do not trigger fairness regularization. These failures often arise when salient object and scene semantics in the top-1 anchor $y^*$ overwhelm demographic-correlated cues, causing the semantic--attribute discrepancy to exceed the threshold.

\section{Conclusion}
We introduced Reward-Gated Test-Time Adaptation (RG-TTA), a selective test-time debiasing framework for CLIP-style vision--language models.
RG-TTA uses an input-dependent reward gate to activate an attribute-balancing term only when necessary, together with episodic updates and parameter resets to limit unintended drift.
Experiments across in-domain and out-of-domain fairness benchmarks as well as standard zero-shot utility tasks show that RG-TTA consistently reduces demographic skew while maintaining competitive utility.
Overall, our results highlight selective, input-conditioned adaptation as a practical design principle for mitigating bias without broadly disrupting model behavior.

\section*{Limitations}
Our approach relies on test-time adaptation (TTA) with per-query parameter updates, which can increase computation and latency, especially when using many update steps or large candidate sets. Because offline-trained debiasing methods amortize cost during training whereas TTA incurs cost online, direct runtime comparisons across these paradigms are not always apples-to-apples and can vary with the deployment scenario. Our experiments focus on a single protected attribute; extending the framework to multiple attributes requires more complex reward/constraint design and may introduce conflicting objectives. The method also assumes access to an external reward signal from a stronger model, which raises availability and cost considerations and may transfer the reward model’s own biases into the adaptation signal. Moreover, our fairness objective implicitly targets proximity to a chosen reference distribution (e.g., uniform), and both evaluation and reward depend on the accuracy and domain robustness of attribute estimators; TTA behavior can further be sensitive to hyperparameters and may be unstable for some queries. Future work will develop more efficient update schemes to reduce online overhead. We will also explore scalable multi-attribute objectives/constraints and robustness techniques to mitigate sensitivity to reward sources and attribute estimators.

\section*{Ethics Statement}
This work proposes a selective test-time adaptation (TTA) approach to mitigate distributional biases over protected attributes (e.g., gender, age, and race) that can arise for person-centric queries. Our experiments use publicly available fairness evaluation datasets (e.g., FairFace, UTKFace, and FACET) together with standard utility benchmarks, and quantify bias using distribution-based fairness metrics. Because face images and protected-attribute annotations can be sensitive, our study does not aim to identify individuals and assumes use strictly in accordance with the datasets' licenses and usage conditions. Our method further relies on an external reward signal from a stronger model (e.g., a fixed CLIP ViT-L/14 reference), which introduces practical considerations about the availability and cost of such signals and raises the possibility that biases present in the reward model could be propagated through the adaptation process. In addition, our fairness objective implicitly assumes a chosen target distribution (e.g., a uniform prior), which may not be appropriate for all tasks or domains. We therefore recommend that any real-world deployment be accompanied by careful auditing of the reward source and attribute estimators for bias and error, and that use in high-stakes decision-making contexts be avoided or subjected to additional, domain-specific validation and oversight.

\section*{Acknowledgments}
This work was partly supported by Institute of Information and 
Communications Technology Planning and Evaluation (IITP) grant funded 
by the Korea Government (MSIT) (No. RS-2022-II220184, Development and 
Study of AI Technologies to Inexpensively Conform to Evolving Policy 
on Ethics), partly supported by the Institute of Information and 
Communications Technology Planning and Evaluation (IITP) grant funded 
by the Korea Government (MSIT) [RS-2021-II211341, Artificial 
Intelligence Graduate School Program (Chung-Ang University)], and 
partly supported by the National Research Foundation of Korea (NRF) 
grant funded by the Korea government (MSIT) [RS-2026-25498346].

\nocite{*}
\bibliography{latex/custom}

\appendix

\section{Implementation Details}
\label{sec:add_impl}
In this section, we provide comprehensive details regarding the experimental setup, optimization hyperparameters, and the construction of the bias subspace, complementing the summary provided in the main text.

\subsection{Computational Environment and Models}
All experiments were conducted on a single \textbf{NVIDIA A6000 GPU} with automatic mixed precision (AMP) enabled to enhance efficiency. We utilized the official pre-trained CLIP checkpoints~\citep{radford2021clip} for ViT-B/16, ViT-B/32, and ViT-L/14 backbones. Consistent with the episodic TTA protocol, the CLIP reward signal was computed using a separate, fixed \textbf{CLIP ViT-L/14} model to provide stable guidance.

\subsection{Modality-Specific Update Strategy}
During the episodic adaptation, we update only the encoder corresponding to the \textit{input query modality} to align it with the frozen target modality:
\begin{itemize}
    \item \textbf{Text-to-Image Retrieval:} We adapt the \textbf{text encoder} while keeping the image encoder frozen.
    \item \textbf{Image-to-Text Retrieval \& Classification:} We adapt the \textbf{image encoder} while keeping the text encoder frozen.
\end{itemize}

\subsection{Optimization Hyperparameters}
We used the \textbf{AdamW} optimizer~\citep{loshchilov2019decoupled} with a learning rate of $1 \times 10^{-4}$ and a weight decay of $0$.
The balancing coefficients for the reward function were set to $\lambda = 1000$ (debiasing penalty weight).
Additionally, we set the Gaussian kernel temperature parameter $\gamma = 0.25$ to control the sharpness of soft assignments in the bias subspace.

\subsection{Adaptive Computational Budgeting}
The Reward Gate (RG) determines the bias sensitivity using a fixed threshold of $\epsilon = 0.02$. Based on the gate's output $\delta(q)$, we dynamically adjust the number of update steps ($T$) and the candidate pool size ($K$):
\begin{itemize}
    \item \textbf{Bias-Insensitive ($\delta(q)=0$):} We perform a lightweight update with $T = 3$ steps and $K = 10$ candidates.
    \item \textbf{Bias-Sensitive ($\delta(q)=1$):} We increase the budget to $T = 10$ steps and $K = 1024$ candidates. In this case, candidates are ranked by their CLIP alignment scores to filter the most relevant samples for the update.
\end{itemize}

% ==============Table5 : Fairface Race성능====================
\begin{table}[t]
  \centering
  \caption{Race debiasing performance using FairFace as the source.}
  \label{tab:src_fairface_race}

  \setlength{\tabcolsep}{3.5pt}
  \renewcommand{\arraystretch}{1.1}

  \resizebox{\columnwidth}{!}{
    % Backbone | Methods | UTKFace(2) | IN1K(2) | Flickr(2) | ABLE(1)
    \begin{tabular}{c|c|cc|cc|cc|c}
    \toprule
    \multirow{3}{*}{\centering Backbone} &
    \multirow{3}{*}{\centering Methods} &
    \multicolumn{2}{c|}{Fairface} &
    \multicolumn{2}{c|}{IN1K} &
    \multicolumn{2}{c|}{Flickr} &
    \multirow{3}{*}{\centering ABLE (\%)$\uparrow$} \\
    
     \cline{3-4} \cline{5-6} \cline{7-8}
    & & & &
    \multicolumn{2}{c|}{\small Acc. (\%)$\uparrow$} &
    \multicolumn{2}{c|}{\small R@5 (\%)$\uparrow$} &
    \\
    
    & &
    {\footnotesize MS$\downarrow$} & {\footnotesize NDKL$\downarrow$} &
    {\footnotesize Top-1} & {\footnotesize Top-5} &
    {\footnotesize TR} & {\footnotesize IR} &
    \\
    \midrule

      % ===================== ViT-B/16 =====================
      \multirow{5}{*}{\centering ViT-B/16}
      & Original CLIP    & 0.528 & 0.182 & 68.31 & 91.83 & 96.4 & 85.5 & 63.31\\
      & CLIP-clip        & 0.544 & 0.161 & 67.97 & 91.62 & 95.4 & 85.3 & 62.62 \\
      & Biased-prompts   & 0.518 & 0.219 & 67.00 & 90.72 & 94.1 & 85.8 & 63.07\\
      & Joint V-L        & \textbf{0.353} & \textbf{0.125} & 68.07 & 91.64 & 96.5 & 83.8 & 69.14 \\
      & Ours             & 0.372 & 0.167 & \textbf{70.32} & \textbf{92.96}
                         & \textbf{97.2} & \textbf{88.4} & \textbf{69.62} \\
      \midrule

      % ===================== ViT-B/32 =====================
      \multirow{5}{*}{\centering ViT-B/32}
      & Original CLIP    & 0.568 & 0.165 & 63.39 & 88.83 & 94.7 & 83.5 & 59.84 \\
      & CLIP-clip        & 0.713 & 0.227 & 62.51 & 88.33 & 92.6 & 81.2 &  54.95\\
      & Biased-prompts   & 0.595 & 0.282 & 61.80 & 87.46 & 91.9 & 83.3 &  58.29\\
      & Joint V-L        & 0.503 & 0.149 & 63.07 & 88.61 & 94.2 & 83.0 & 61.74 \\
      & Ours             & \textbf{0.389} & \textbf{0.148} & \textbf{69.73} & \textbf{92.26}
                         & \textbf{97.0} & \textbf{86.5} & \textbf{68.74} \\
      \midrule
    \end{tabular}
  }
\end{table}

\subsection{Bias Subspace Construction Details}
The bias subspace is constructed offline using the training split of the dataset:
\begin{itemize}
    \item \textbf{Text Queries:} We use attribute-specific prompts formatted as ``\textit{a photo of a \{attribute class\} person}''. One prompt is generated per attribute class.
    \item \textbf{Image Queries:} We construct the reference set $A$ by sampling $M = 5$ images per attribute class uniformly from the \textbf{UTKFace} dataset. To ensure reproducibility and consistency, this reference set is fixed once and reused across all test episodes. The total size of the reference set is $|A| = C \times 5$, where $C$ is the number of attribute classes.
\end{itemize}

% \section{Additional Experiments}
\section{Additional Results}
\subsection{Additional Race Debiasing Results using FairFace}
\label{app:race_fairface}

In the main text (Table~\ref{tab:src_utkface_race}), we utilized UTKFace as the source dataset for constructing the bias subspace to mitigate Race bias. To verify the robustness of our framework across different source domains, we conducted an additional experiment using \textbf{FairFace} as the source dataset.

The results are presented in Table~\ref{tab:src_fairface_race}. Consistent with the findings in the main text, our RG-TTA framework demonstrates a superior capability to balance fairness and utility. Regarding fairness, our method effectively mitigates racial bias compared to the Original CLIP; while \textbf{Joint V-L} shows competitive scores on ViT-B/16, our method achieves the best performance on ViT-B/32 (MaxSkew: 0.389). Crucially, in terms of utility preservation, our method consistently outperforms all baselines in zero-shot tasks (ImageNet and Flickr) across both backbones, confirming that our selective routing mechanism successfully prevents the over-debiasing observed in static approaches. Consequently, our method achieves the highest ABLE scores for both backbones (69.62\% and 68.74\%), proving that it maintains the optimal trade-off between fairness and utility regardless of the source dataset used.

\begin{figure}[t]
    \centering
    \setlength{\fboxsep}{0pt}

    \begin{subfigure}[t]{0.49\linewidth}
        \centering
        \fbox{\includegraphics[
            height=4cm,
            keepaspectratio
        ]{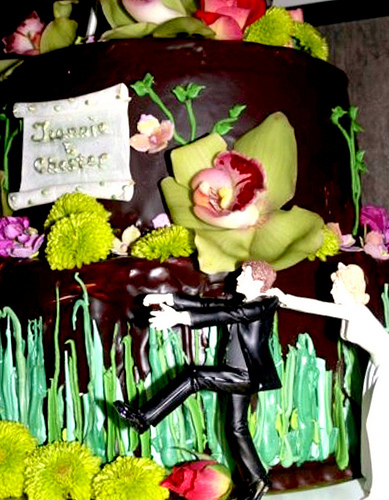}}
    \end{subfigure}
    \begin{subfigure}[t]{0.49\linewidth}
        \centering
        \fbox{\includegraphics[
            height=4cm,
            keepaspectratio
        ]{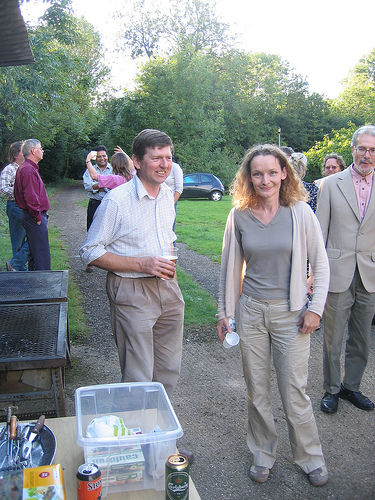}}
    \end{subfigure}

\caption{\textbf{Gating failure cases on ImageNet.} In ImageNet, queries are expected to be bias-insensitive and thus gated off ($\delta(q)=0$). Shown are false-positive cases where the gate is incorrectly activated ($\delta(q)=1$), causing debiasing and alignment rewards to be jointly applied.}

    \label{fig:imagenet_gating_failures}
\end{figure}

\subsection{False-positive gating failures on ImageNet}

Figure~\ref{fig:imagenet_gating_failures} illustrates \textit{false-positive} cases on ImageNet-1K, where bias-insensitive object queries are incorrectly gated on ($\delta(q)=1$), triggering unnecessary fairness regularization. These failures typically arise when the queried object strongly co-occurs with humans or human-like features in the top-1 retrieved anchor $y^*$. For instance, as shown in Figure~\ref{fig:imagenet_gating_failures}, human figurines on a cake (left) or bystanders in the background (right) provide strong demographic signals that reduce the semantic-attribute discrepancy below the threshold $\epsilon$. This misleads the gate into treating the object query as bias-sensitive. Although our hybrid reward design minimizes semantic drift even when the gate is mistakenly active, these cases represent a computational inefficiency.

\subsection{Attribute-Specified Queries}
\label{app:acr}

A natural question is whether the gating mechanism may overreact to queries that explicitly specify a demographic attribute (e.g., ``a photo of a male doctor''), thereby distorting query intent. Since such queries are likely to align with the corresponding attribute prompts in $\mathcal{A}$, the gate may activate $\delta(q){=}1$ and apply the attribute-balancing reward, which could in principle override the user's specified attribute. To quantitatively assess this behavior, we evaluate on FACET (ViT-B/16) using gender-specified queries. We introduce the \emph{Attribute Consistency Rate} (ACR@$k$): the proportion of top-$k$ retrieved results whose gender matches the query-specified gender. Higher values indicate better preservation of query intent.

As expected, gender-specified queries trigger $\delta(q){=}1$, since they are close to the corresponding attribute prompts. Nevertheless, RG-TTA limits ACR degradation to only 1.5--2.6 percentage points compared to Original CLIP, because the CLIP alignment reward $r_{\text{clip}}$ in Eq.~5 assigns higher scores to gender-matching candidates, effectively counterbalancing the attribute-balancing penalty. To verify this, we evaluate a \emph{Debiasing Only} variant that removes $r_{\text{clip}}$ and applies the attribute-balancing reward alone: ACR@10 drops sharply by 0.16 (0.692$\rightarrow$0.532), confirming that the CLIP reward is the key component preventing this failure mode.

\subsection{Intersectional Debiasing}
\label{app:intersectional}

Real-world fairness concerns often involve multiple protected attributes simultaneously (e.g., gender and race). RG-TTA accommodates intersectional settings by defining the class set $\mathcal{C}$ in Eq.~3--4 as a Cartesian product of attributes (e.g., gender $\times$ race $\rightarrow$ \{male-White, male-Black, $\ldots$\}), without modifying the pipeline. Rather than introducing separate reward terms per attribute (which can cause multi-objective conflicts), this consolidates multi-attribute balancing into a single reward over cross-product classes.

\begin{table}[t]
\centering
\small
\caption{Attribute Consistency Rate (ACR@$k$) for gender-specified queries on FACET (ViT-B/16). ``Debiasing Only'' removes the CLIP alignment reward and forces $\delta(q){=}1$.}
\label{tab:acr}
\begin{tabular}{lccc}
\toprule
Method & ACR@10 & ACR@25 & ACR@50 \\
\midrule
Original CLIP   & 0.692 & 0.666 & 0.649 \\
RG-TTA (Ours)   & 0.672 & 0.651 & 0.623 \\
Debiasing Only  & 0.532 & 0.524 & 0.513 \\
\bottomrule
\end{tabular}
\end{table}

\begin{table}[t]
\centering
\small
\caption{Intersectional debiasing on UTKFace (gender $\times$ race, ViT-B/16).}
\label{tab:intersectional}
\begin{tabular}{llcc}
\toprule
Attribute & Method & MaxSkew$\downarrow$ & NDKL$\downarrow$ \\
\midrule
\multirow{2}{*}{Gender}
 & Original CLIP & 0.114 & 0.080 \\
 & RG-TTA (Ours) & 0.072 & 0.053 \\
\midrule
\multirow{2}{*}{Race}
 & Original CLIP & 0.575 & 0.137 \\
 & RG-TTA (Ours) & 0.381 & 0.089 \\
\bottomrule
\end{tabular}
\end{table}

As shown in Table~\ref{tab:intersectional}, RG-TTA reduces Gender MaxSkew by 36.8\% and Race MaxSkew by 33.7\% \emph{simultaneously}, confirming effective multi-attribute balancing. As a test-time method, only the class definition needs to change—no retraining is required. We acknowledge that exponential growth of cross-product classes (e.g., 2$\times$4$\times$3=24 for three attributes) may degrade the popularity estimate $p_c$, and a more scalable formulation for high-dimensional intersectional settings remains an open direction.

\subsection{Generalization to BLIP}
\label{app:blip}

To verify that RG-TTA generalizes beyond CLIP to other contrastive vision-language models, we apply it to BLIP-base (\texttt{Salesforce/blip-itm-base-coco}) using its image-text contrastive (ITC) branch. We evaluate on UTKFace with gender as the protected attribute.

\begin{table}[t]
\centering
\small
\caption{RG-TTA applied to BLIP-base on UTKFace (Gender). RG-TTA generalizes to BLIP without retraining; only the gating threshold $\epsilon$ requires recalibration.}
\label{tab:blip}
\begin{tabular}{lcc}
\toprule
Method & MaxSkew$\downarrow$ & NDKL$\downarrow$ \\
\midrule
Original BLIP & 0.236 & 0.115 \\
RG-TTA (Ours) & 0.106 & 0.044 \\
\bottomrule
\end{tabular}
\end{table}

RG-TTA achieves a 55.1\% reduction in MaxSkew and a 61.7\% reduction in NDKL on BLIP, confirming applicability to contrastive VLMs beyond CLIP. Notably, no retraining is required; the entire framework operates at test time.

\paragraph{Threshold recalibration.} 
Different architectures produce different similarity score distributions, so the gating threshold $\epsilon$ does not transfer directly across models. Applying CLIP's default $\epsilon{=}0.02$ to BLIP yields 86.2\% $\delta(q){=}1$ on UTKFace and 8.1\% on ImageNet (compared to 99.9\% / 0.1\% on CLIP). However, recalibration only involves adjusting a single scalar with a small validation set—no architectural changes or retraining are needed.

\subsection{Generalization Beyond Demographic Bias}
\label{app:waterbirds}

While RG-TTA primarily targets social fairness with demographic attributes, the underlying mechanism—selective debiasing in a bias subspace—is in principle applicable to other types of distributional bias. To briefly probe this generalization, we apply RG-TTA to the Waterbirds benchmark, where bias arises from spurious correlations between bird species and background scenes (water vs.\ land), rather than from protected demographic attributes. We set the background as the bias subspace attribute and measure the distribution bias in top retrieval results for neutral queries (e.g., ``a photo of a bird'').

\begin{table}[t]
\centering
\small
\caption{RG-TTA on Waterbirds (ViT-B/16). The bias subspace attribute is set to background (water vs.\ land).}
\label{tab:waterbirds}
\begin{tabular}{lcc}
\toprule
Method & MaxSkew$\downarrow$ & NDKL$\downarrow$ \\
\midrule
Zero-shot CLIP & 0.239 & 0.134 \\
RG-TTA (Ours)  & 0.186 & 0.070 \\
\bottomrule
\end{tabular}
\end{table}

By simply redefining the bias subspace attribute (demographic $\rightarrow$ background), RG-TTA reduces MaxSkew by 22.2\% and NDKL by 47.8\% without retraining. While a thorough investigation in spurious-correlation settings is beyond the scope of this work, this result suggests broader applicability of the proposed selective adaptation principle.

% \section{Quailtative Analysis}
% \subsection{Bias Mitigation}
% \subsection{Utility Preservation}
% \subsection{Failure case analysis in Imagenet dataset}

% \subsection{Runtime and Compute Overhead}
% δ=0 vs δ=1 latency, FLOPs, memory

% 민감 쿼리 비율 분포

% ---------------------------------------------------------
% Example wide figure placeholder
% ---------------------------------------------------------

% ACL은 Limitations 섹션을 요구하는 경우가 많아서 유지
% \paragraph{Page budget (rough).} About 0.2--0.4 pages.

% \section*{Acknowledgments}
% (Optional) Keep short.

% =======================
% References
% =======================
% custom.bib에 있는 엔트리를 참고문헌에 넣기 위한 예시

% =======================
% Knob: page-count tuning
% =======================
% 만약 컴파일 결과가 9페이지보다 짧으면 범위를 늘리세요 (예: [69-75]).
% 너무 길면 범위를 줄이세요.
% \EXTRAFILL  % 필요할 때 주석 해제

\end{document}